\documentclass[10pt,twocolumn,letterpaper]{article}

\usepackage[pagenumbers]{iccv} 

\definecolor{iccvblue}{rgb}{0.21,0.49,0.74}
\usepackage[pagebackref,breaklinks,colorlinks,allcolors=iccvblue]{hyperref}

\usepackage{url}            
\usepackage{booktabs}       
\usepackage{amsfonts}       
\usepackage{nicefrac}       
\usepackage{microtype}      
\usepackage{amsmath}
\usepackage{multirow}
\usepackage{bm}
\usepackage{paralist} 
\usepackage{graphicx}
\usepackage{wrapfig}
\usepackage{bbding}
\usepackage{color}
\usepackage{cite}
\usepackage{amsthm,amsmath,amssymb}
\usepackage{mathrsfs}
\usepackage{stfloats}
\usepackage{amssymb}
\usepackage{adjustbox}

\usepackage[percent]{overpic}
\usepackage{enumitem}
\usepackage{relsize}
\usepackage{xcolor}
\usepackage{booktabs}
\usepackage{colortbl}

\def\ie{i.e.} 

\def\bes#1{{\color{black}\underline{#1}}}
\def\cite{\citep}

\definecolor{lightblue}{RGB}{173,216,230}
\definecolor{mydarkred}{RGB}{139,0,0}
\definecolor{c1}{HTML}{d6ecf0}

\usepackage{float}

\def\paperID{5817} 
\def\confName{ICCV}
\def\confYear{2025}

\begin{document}

\title{TAViS: Text-bridged Audio-Visual Segmentation with Foundation Models}

\author{
	Ziyang Luo$^{1}$
	\hspace{5pt}
	Nian Liu$^{2}$
	\hspace{5pt}
	Xuguang Yang$^{1}$
	\hspace{5pt}
        Salman Khan$^{2}$
        \hspace{5pt}
        Rao Muhammad Anwer$^{2}$\\
        \hspace{5pt}
        Hisham Cholakkal$^{2}$
        \hspace{5pt}
        Fahad Shahbaz Khan$^{2}$
        \hspace{5pt}
        Junwei Han$^{1,3}$
	\\
	$^1$Northwestern Polytechnical University
	$^2$Mohamed bin Zayed University of Artificial Intelligence\\
    $^3$ Institute of Artificial Intelligence, Hefei Comprehensive National Science Center 
}

\twocolumn[{%
    \maketitle
    \begin{figure}[H]
        \hsize=\textwidth  
        \centering
        \includegraphics[width=1\textwidth]{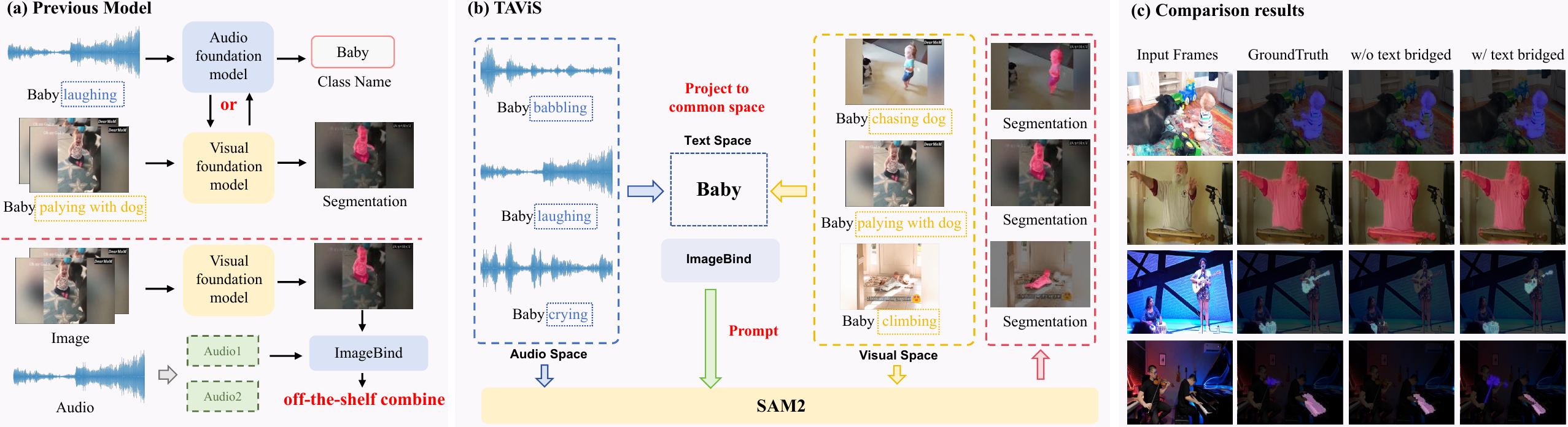}
        \caption{\textbf{Illustration of TAViS compared with previous methods.}
        (1) Previous methods often rely on single-modality foundation models (top) or combine the visual foundation model with ImageBind in an off-the-shelf manner (bottom), which limits their ability to address the misalignment of audio-visual information with significant intra-class diversity.
        (2) We propose TAViS, a novel framework that leverages two foundation models: ImageBind and SAM2. The text-bridged alignment between visual and audio modalities serves as both prompts and supervision signals for TAViS.
        (3) We demonstrate the effectiveness of our approach through comparative experiments with and without the text-bridged mechanism.}
        \label{compare_with_previous}
    \end{figure}
}]

\begin{abstract}
Audio-Visual Segmentation (AVS) faces a fundamental challenge of effectively aligning audio and visual modalities. While recent approaches leverage foundation models to address data scarcity, they often rely on single-modality knowledge or combine foundation models in an off-the-shelf manner, failing to address the cross-modal alignment challenge. In this paper, we present TAViS, a novel framework that \textbf{couples} the knowledge of multimodal foundation models (ImageBind) for cross-modal alignment and a segmentation foundation model (SAM2) for precise segmentation. However, effectively combining these models poses two key challenges: the difficulty in transferring the knowledge between SAM2 and ImageBind due to their different feature spaces, and the insufficiency of using only segmentation loss for supervision. To address these challenges, we introduce a text-bridged design with two key components: (1) a text-bridged hybrid prompting mechanism where pseudo text provides class prototype information while retaining modality-specific details from both audio and visual inputs, and (2) an alignment supervision strategy that leverages text as a bridge to align shared semantic concepts within audio-visual modalities. Our approach achieves superior performance on single-source, multi-source, semantic datasets, and excels in zero-shot settings. 
\end{abstract} 
\section{Introduction}
\label{sec:intro}
Audio-visual joint perception ability is crucial for our understanding of the surrounding environment.
Various audio-visual collaboration tasks have emerged in recent years, including audio-visual sound separation \cite{gao2021visualvoice,tzinis2022audioscopev2}, visual sound source localization \cite{chen2021localizing,hu2020discriminative,hu2021class,qian2020multiple}, and audio-visual video understanding \cite{kazakos2019epic,lee2020cross,lin2019dual}. Distinct from these tasks, audio-visual segmentation (AVS) \cite{zhou2022audio, zhou2024audio} aims to generate pixel-wise segmentation maps of sounding objects within a scene, which has garnered increasing attention from the research community.

The primary challenge in AVS lies in effectively aligning audio and visual modalities, which requires capturing complex relationships between audio-visual patterns across diverse scenarios.
Previous approaches are constrained by limited training data, resulting in incomplete capture of modal relationships. To mitigate this limitation, researchers have introduced large-scale datasets such as AVS-Synthetic \cite{liu2024annotation} and VPO \cite{chen2024unraveling}. While pre-training on these datasets has shown promising improvements, it significantly increases computational overhead during training. Thanks to the generalization capabilities of foundation models, several works \cite{yang2024cooperation,mo2023av,liu2024annotation, liu2024bavs} have utilized image foundation models, such as semantic-SAM \cite{li2023semantic} or SAM \cite{chen2023sam}, as well as audio foundation models (e.g., Beats \cite{chen2022beats}), to leverage pre-existing knowledge with minimal additional training on AVS datasets. However, these approaches typically rely on single-modality foundation knowledge, either from audio or image, which hinders effective audio-visual modality alignment. 
While some previous works \cite{bhosale2023leveraging, bhosale2024unsupervised} attempted to combine segmentation and multimodal foundation models, they employed these models in an off-the-shelf manner for unsupervised settings, where the models operated independently without meaningful interaction, as shown in Figure~\ref{compare_with_previous} (a). 
This reliance on a decoupled combination strategy compromises the transfer of foundation model knowledge.

In this paper, we propose a new model TAViS as shown in Figure~\ref{compare_with_previous}  (b), designed to couple the knowledge provided by foundation models as a straightforward yet effective solution for the AVS task.  However, choosing foundation models for the AVS task is challenging, as it requires both accurate recognition of sound-making objects and detailed segmentation of their boundaries.
Our analysis leads us to focus on two complementary foundation models: ImageBind \cite{girdhar2023imagebind} for its robust audio-visual alignment, and SAM2 \cite{ravi2024sam} for its strong generalization capability in complex image and video segmentation. To adapt SAM2 for object-level segmentation from mixed audio guidance, we design an ImageBind-guided query decomposition block (IBQD) that effectively separates audio sources while maintaining ImageBind's well-aligned audio feature space.

After selecting these foundation models, we identify two main challenges in their integration for the AVS task. First, directly transferring ImageBind's alignment knowledge to SAM2's segmentation framework is difficult due to their different feature spaces. Second, using only segmentation loss for supervision is insufficient as it merely implies the need for alignment knowledge.
To address the first challenge, we design a text-bridged hybrid prompting technique where text provides class prototype information while audio and visual modalities complement with modality-specific details. 
For the second challenge of alignment supervision, we propose text-bridged alignment supervision with two separate losses: audio-to-text and image-to-text. While audio and visual modalities contain object-specific information (e.g., different timbre and backgrounds for the same class),
text can help extract and align their shared semantic concepts, providing a more effective supervision signal than direct audio-visual alignment.
Furthermore, text enables a more generalized classification approach by leveraging similarity comparisons between textual and audio-visual embeddings, rather than relying on a fixed MLP classifier. Based on this similarity-based approach, we further enable TAViS with zero-shot generalization capabilities.

Our main contributions are summarized as follows:
\begin{compactitem}
\item We \textbf{couple} complementary strengths of foundation models for AVS, fusing ImageBind's cross-modal understanding with SAM2's segmentation precision instead of relying on off-the-shelf combined solutions.
\item We propose a text-bridged hybrid prompting technique, transforming audio queries into pseudo-text embeddings while preserving audio-specific cues, and complementing with ImageBind's image \emph{cls} tokens to align image and audio-text modalities.
\item We design text-bridged alignment supervision to use text as an intermediate bridge for better aligning shared semantic concepts across modalities.
\item Our model demonstrates superior performance across diverse AVS benchmarks while excelling in zero-shot settings, establishing a versatile framework that unifies binary, semantic, and zero-shot segmentation capabilities within a single architecture.
\end{compactitem}

\vspace{-3mm}
\section{Related Work}
\label{sec:relatedwork}
\subsection{Audio-Visual Segmentation}
In recent years, AVS \cite{zhou2023audio} task has garnered significant attention, as it addresses the challenging problem of locating and predicting pixel-wise maps for sounding objects within a scene. Existing AVS methods can be broadly categorized into three approaches: fusion-based, generative-based, and foundation model-based methods. Fusion-based methods constructed relationships between audio and visual inputs using unidirectional or bidirectional cross-attention mechanisms  \cite{zhou2023audio,li2023catr,gao2024avsegformer,li2024qdformer,yang2024cooperation, ling2024transavs}. For generative-based methods, latent diffusion models \cite{mao2023contrastive} or variational auto-encoders \cite{mao2023multimodal} were utilized to achieve accurate object segmentation. Foundation model-based methods leveraged visual foundation models such as SAM \cite{kirillov2023segment,mo2023av,liu2024annotation,nguyen2024save, seon2024extending}, UniDiffuser \cite{liu2024bavs} and SAM-semantic \cite{li2023semantic} to obtain enhanced visual knowledge, facilitating the establishment of audio-visual relationships. However, these aforementioned foundation model-based methods primarily focus on visual prior knowledge, leaving a domain gap between audio and visual modalities unaddressed.

In this paper, we propose a novel approach that combines the strengths of Imagebind \cite{girdhar2023imagebind} and SAM2 \cite{ravi2024sam}. Unlike BAVS \cite{liu2024bavs}, which utilizes text to align prediction maps, our method designs two separate losses to align text with visual and audio respectively, fully leveraging the text modality for cross-modal alignment. While \cite{bhosale2023leveraging, bhosale2024unsupervised} also combined SAM and ImageBind for unsupervised audio-visual segmentation, they merely integrated these models in an off-the-shelf manner, which means the two models operate independently without interaction. In contrast, we introduce a text-bridged hybrid prompting design to align ImageBind with SAM2, thereby creating an organically coupled system that leads to more effective optimization. 


\begin{figure*}[!t]
    \centering
    \includegraphics[width=1\linewidth]{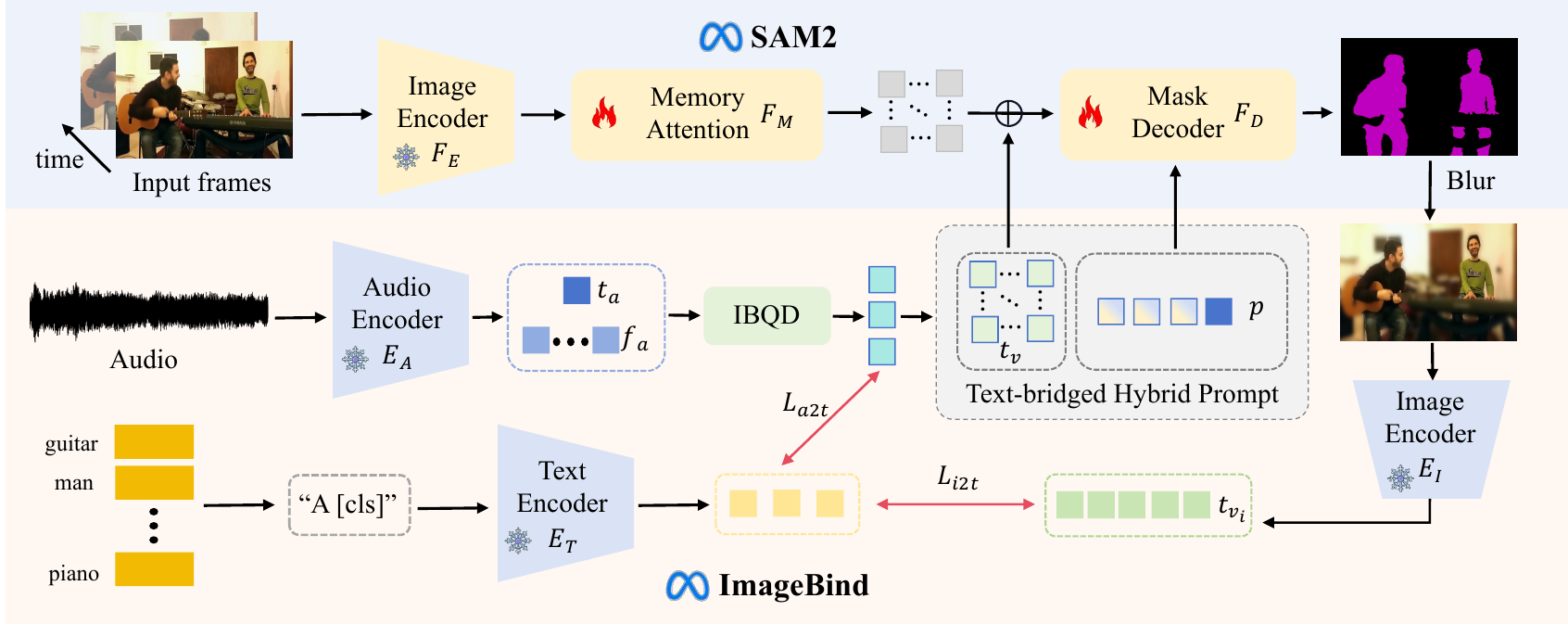}
    \caption{\textbf{Overall framework of our proposed model.} 
    Our framework integrates two foundation models: SAM2 for precise segmentation and ImageBind for audio-visual alignment. The input audio is first processed by the audio encoder to extract audio embeddings, which are then decomposed into object-level queries via the IBQD module. Meanwhile, the input image is processed by SAM2. The text-bridged hybrid prompts, including sparse prompt ($\bm{p}$) and dense prompt ($\bm{t}_v$), are then generated and fed into the mask decoder to obtain segmentation masks. Finally, two text-bridged alignment losses $\mathcal{L}_{{a2t}}$ and $\mathcal{L}_{{i2t}}$ are applied to supervise the audio-to-text and image-to-text relationships, respectively.}
    \label{all}
    \vspace{-0.5cm}
\end{figure*}

\subsection{Multimodal Joint Representation for Audio}
Multimodal joint representation primarily focuses on mapping inputs from different modalities into a shared embedding space. In this section, we specifically focus on audio-related methods.
For dual-modality alignment, WavCaps \cite{mei2024wavcaps} pioneered multimodal representation learning for audio-text pairs, utilizing a large-scale weakly-labeled audio captioning dataset. In the multi-modal context, AudioCLIP \cite{mahmud2023ave} and WAV2CLIP \cite{wu2022wav2clip} extended CLIP's architecture by introducing dedicated audio encoders trained through contrastive learning, leveraging audio-image-text and audio-image pairs respectively. More recent approaches, including ImageBind \cite{girdhar2023imagebind}, LanguageBind \cite{zhu2023languagebind}, and OmniBind \cite{wang2024omnibind}, proposed comprehensive frameworks for integrating multiple modalities by utilizing various data pairs that share common image, language, and text modalities.

Existing AVS methods \cite{yang2024cooperation, mo2023av, liu2024annotation, liu2024bavs} typically rely on single-modality foundation models for either audio or visual processing independently, resulting in heterogeneous feature spaces that complicate cross-modal alignment. While \cite{mo2024unveiling} trained a specialized foundation model for audio-visual tasks, it lacks text alignment capabilities and predicts fixed classes via MLP prediction, which constrains its applicability in real-world scenarios. In contrast to these approaches, our model can perform zero-shot setting, demonstrating superior generalizability for unseen classes through our innovative text-bridged design approach. 

\section{Methodology}
\label{sec:method}
In this work, we propose TAViS, a novel framework that leverages foundation models to achieve precise region segmentation by bridging audio and visual content through text representations. Built upon SAM2 \cite{ravi2024sam} and ImageBind \cite{han2023imagebind} (Section~\ref{sec:sam2+imagebind}), our approach introduces three key components: 1) an ImageBind-guided query decomposition module to decompose the mixed audio embedding obtained from ImageBind into object-level queries (Section~\ref{sec:query}).

2) a text-bridged hybrid prompting mechanism that consists of pseudo-text embeddings and audio queries for sparse prompt and  image \emph{cls} token for dense prompt  (Section~\ref{sec:prompt}).
And 3) two text-bridged alignment loss functions, $\mathcal{L}_{{a2t}}$ and $\mathcal{L}_{{i2t}}$, which establish robust audio-visual relationships using text as an intermediate bridge (Section~\ref{sec:loss}). The overall architecture of our framework is illustrated in Figure~\ref{all}.

\subsection{Preliminaries}
\label{sec:sam2+imagebind}
For visual segmentation, we adopt SAM2 \cite{ravi2024sam} as our foundation model. SAM2 consists of three key components: an image encoder $F_E$ to extract visual features, a memory attention module $F_M$ to maintain temporal consistency, and a mask decoder $F_D$ to generate precise segmentation masks for regions of interest.
To establish effective audio-visual alignment, we incorporate ImageBind \cite{han2023imagebind} into our framework. Within the ImageBind architecture, we specifically utilize its image encoder $E_I$, text encoder $E_T$, and audio encoder $E_A$.
For each modality encoder, we obtain the trunk feature and a \emph{cls} token with a transformer architecture. 

Given an audio signal $\bm{A}$ and its corresponding video frames $\bm{I} \in{\mathbb{R}^{T\times H \times W \times C}}$, where $T$ denotes the total number of frames in the video, we first extract visual features through SAM2's image encoder $F_E$, generating multi-stage visual features.
Concurrently, we process the audio input $\bm{A}$ through ImageBind's audio encoder $E_A$ to obtain the audio trunk feature $\bm{f}_a$ and the \emph{cls} token $\bm{t}_a$. To enhance the correlation between audio and visual modalities, we introduce audio-guided adaptors at each stage of SAM2's visual encoder $F_E$, following the approach in \cite{liu2024annotation}.

\subsection{ImageBind-guided Query Decomposition}
\label{sec:query}
Previous SAM-based AVS methods \cite{kirillov2023segment,mo2023av,liu2024annotation,nguyen2024save} typically treat the entire audio feature as a single prompt for the mask decoder. However, the mixing of information from multiple sound sources conflicts with the object-specific queries required by SAM2. A straightforward way to mitigate this issue is to decompose the entire audio feature to object-level prompts with a learnable decomposition module. However, naively doing this can not leverage the well-aligned multimodal knowledge learned by ImageBind. To address this challenge, we propose an ImageBind-guided Query Decomposition (IBQD) to decompose audio features into object-specific queries that preserve the aligned feature space of ImageBind, enabling more precise subsequent audio-visual correspondence.

For the audio trunk feature $\bm{f}_a \in{\mathbb{R}^{F \times C}}$ and audio \emph{cls} token  $\bm{t}_a \in{\mathbb{R}^{1 \times C}}$ obtained from audio encoder $\bm{E}_A$,  we introduce learnable queries $\bm{t}_W \in{\mathbb{R}^{N \times C}} $ to generate decomposed audio sources. Here, $F$ represents the temporal dimension, $C$ is the channel dimension which represents different frequencies, and $N$ indicates the number of learnable queries. To effectively decompose $\bm{f}_a$ along the frequency dimension,
we employ multi-head cross attention (MCA) mechanisms and utilize weight matrices $W_q$, $W_k$, and $W_v$ to generate the query $Q$, key $K$, and value $V$, as follows:
\begin{equation} \label{query_NMF}
\begin{split}
\bm{t}_W &= \text{Softmax}((W_q\bm{t}_W)(W_k\bm{f}_a^T))(W_v\bm{f}_a) + \bm{t}_W.
\end{split}
\end{equation}
For simplicity, we omit the layer normalization \cite{ba2016layer} operation here. Subsequently,  a linear transformation is applied to $\bm{t}_W$ to generate object-specific biases, which are added to $\bm{t}_a$ and obtain $\bm{t}'_a\in {\mathbb{R}^{N \times C}}$. Finally, the $\bm{t}'_a$ is updated by $\bm{f}_a$ again with MCA: 
\begin{equation} \label{query}
\begin{split}
\bm{t}'_a &= \bm{t}_a + \text{Linear}(\bm{t}_W),\\
\bm{t}'_a &= \text{Softmax}((W_q\bm{t}'_a)(W_k\bm{f}_a^T))(W_v\bm{f}_a) + \bm{t}'_a.
\end{split}
\end{equation}
The decomposed queries $\bm{t}'_a$ in our IBQD module accurately target individual sound sources, aligning with SAM2’s object-centric approach. 
Our decomposed queries are generated as the sum of the original \emph{cls} token $\bm{t}_a $ and biases, thereby preserving the alignment audio feature space of $\bm{t}_a $ and preventing potential disruptions in subsequent alignment.

\begin{figure}[!t]
    \centering
    \includegraphics[width=1\linewidth]{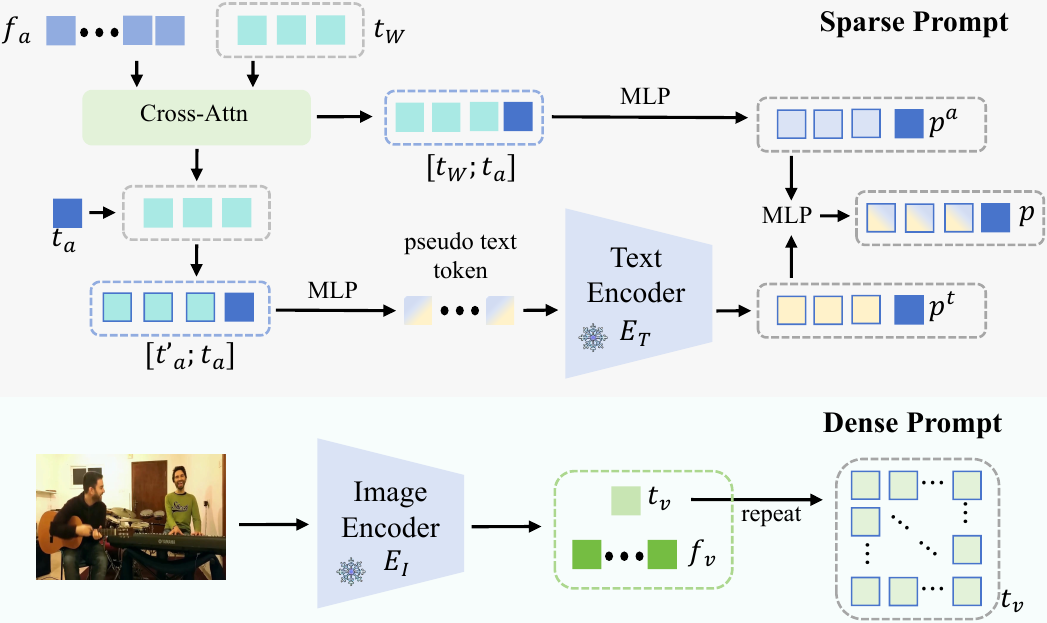}
    \caption{\textbf{Illustration of our text-bridged hybrid prompting mechanism.} For the sparse prompt, we combine pseudo text embeddings and specific audio queries generated through MLP. For the dense prompt, we process the image input through ImageBind to obtain the \emph{cls} token, which is repeated and applied across all pixel locations in SAM2's image feature. 
    }
    \label{prompt_FIG}
    \vspace{-0.5cm}
\end{figure} 

\subsection{Text-bridged Hybrid Prompting}
\label{sec:prompt}
Prompts play a crucial role in SAM2, providing essential guidance to specify the object of interest to segment. 
Previous works based on SAM \cite{kirillov2023segment,mo2023av,liu2024annotation,nguyen2024save} 
typically process audio inputs through learnable MLPs to generate semantic prompts. However, they may struggle to capture the inherent class prototype nature of audio information required for effective AVS alignment. In AVS scenarios, the initial audio input contains substantial non-semantic information (such as diverse timbres and tones) that can confuse the model. Therefore, we propose novel text-bridged hybrid prompts including sparse prompt and dense prompt, where text provides class prototype information while audio and visual modalities complement with modality-specific details. The overall architecture is shown in Figure~\ref{prompt_FIG}.
\vspace{-2mm}
\paragraph{Sparse Prompt.}
For sparse prompt, we adopt an audio-text dual prompt that leverages text and audio to provide the general object concepts and specific instance details, respectively.
We first generate specific object-level text prototype information and scene-level global context following \cite{wu2023aligning}. Specifically, we utilize learnable MLP to the decomposed audio queries $\bm{t}'_a$ and the audio \emph{cls} token $\bm{t}_a$. These two transformed tokens are then concatenated and processed by the text encoder $E_T$ to generate the pseudo text prompt $\bm{p}^t$.

While the pseudo-text prompt captures high-level prototype information, we need to preserve detailed audio characteristics as well. Therefore, we generate specific audio prompt and global audio context. We concatenate and transform  $\bm{t}_W$ and $\bm{t}_a$ through another MLP to form the complete audio prompt $\bm{p}^a$.
Finally, we aggregate the $\bm{p}^t$ and $\bm{p}^a$ to get the final sparse prompt $\bm{p}$.
The aggregation of these two types of prompts is formulated as follows:
\begin{equation} \label{all_prompt}
\begin{split}
\bm{p}^t &= E_T(\text{MLP}[\bm{t}'_{a}; \bm{t}_{a}]),\\
\bm{p}^a &= \text{MLP}[\bm{t}_W; \bm{t}_a],\\
\bm{p} &= \text{MLP}([\bm{p}^a; \bm{p}^t]),
\end{split}
\end{equation}
where [;] denotes concatenation operation,  and $\bm{p}$ is the final spare prompt for SAM2.

To supervise the text embeddings generated from audio, we utilize the trunk feature $E_T(t^t)$ obtained from ImageBind, which helps predict meaningful text information from audio. Here, $t^t$ denotes the ground truth text label with the template ``A [cls]''. The audio-to-pseudo-text loss $\mathcal{L}_{a2pt}$ is computed as:
\begin{equation} \label{i2t}
\mathcal{L}_{a2pt} = \text{MSE}(E_T(\text{MLP}(\bm{t}'_{a})), E_T(t^t)),
\end{equation}
where MSE denotes the Mean Squared Error loss. 

\vspace{-2mm}
\paragraph{Dense Prompt.}
Since SAM2's original image feature is not aligned with our audio-text dual prompt, we propose to leverage the image representation from ImageBind as dense prompt to supplement aligned visual knowledge. Specifically, we input the video frame $\bm{I}$ into ImageBind's image encoder $E_I$, and extract the \emph{cls} token $\bm{t}_v$ to represent the global visual information. After that, we repeat the image token $\bm{t}_v$ and add it to each pixel location in the SAM2's image embedding.

The sparse prompt effectively transforms the audio information into a text format, which allows us to better capture the core object information while preserving the original audio signal to encode specific semantic characteristics.
As for the dense prompt, the visual token provides the contextual information for the frame feature, which benefits the subsequent mask decoding process with the alignment between the image and audio-text modalities.
Our text-bridged hybrid prompting strategy leverages the complementary strengths of both the sparse and dense prompts for more effective segmentation.

\subsection{Text-bridged Alignment Supervision}
\label{sec:loss}
While SAM2 and ImageBind demonstrate exceptional capabilities in segmentation and modality alignment respectively, establishing robust audio-visual association between them remains challenging. To leverage these pre-trained models, a straightforward approach would be directly inputting audio and image into the ImageBind encoder for feature alignment. However, this proves suboptimal and hard to optimize as audio contains non-semantic information (e.g., timbre and intonation), while images contain diverse context and appearance information. Both may cause significant intra-class diversity. Moreover, relying solely on segmentation loss for supervision proves insufficient, as it merely implies the need for alignment knowledge without explicitly guiding the model to learn meaningful audio-visual association.
Therefore, we propose using text as a ``bridge'' since it can concisely express high-level prototype information. Our text-bridged alignment supervision consists of two loss terms: $\mathcal{L}_{a2t}$ and $\mathcal{L}_{i2t}$, which establish audio-to-text and image-to-text relationships, respectively.

\vspace{-2mm}
\paragraph{Audio-to-Text Loss}
For the audio-to-text loss $\mathcal{L}_{{a2t}}$, we first format the text input as ``A [cls]'', where [cls] represents the class labels. Given that the AVS dataset contains $P$ classes (including the background class), we process these $P$ ground truth text inputs $t^t$ through ImageBind's text encoder $E_T$ and projection layer to obtain text embeddings $\text{proj}(E_T(t^t_k))$, where $k \in {0,1,...,P}$.

Subsequently, the audio queries $t'_a$ are passed through ImageBind's final linear projection layer to obtain their representations $\text{proj}(t'_{a_i})$ in the common embedding space, where $i \in {0,1,...,N-1}$.
Next, we employ the Hungarian matching algorithm to establish optimal assignment between $\text{proj}(E_T(t^t_k))$ and $\text{proj}(t'_{a_i})$.
The audio-to-text loss $\mathcal{L}_{{a2t}}$ is then computed as:
\begin{equation} \label{a2t}
\begin{split}
T_a &= [\text{proj}(t'_{a_0}),\text{proj}(t'_{a_1}),...\text{proj}(t'_{a_{N-1}})],\\
T^t &= [\text{proj}(E_T(t^t_0)),\text{proj}(E_T(t^t_1)),...\text{proj}(E_T(t^t_P))],\\
a2t &= T_a  \times T^t,\\
\mathcal{L}_{{a2t}} &= \text{CELoss}(a2t, y),
\end{split}
\end{equation}
where CELoss denotes the cross-entropy loss, $a2t \in \mathbb{R}^{N \times P}$, and $y \in \mathbb{R}^{N}$ represents the index vector of the matched class labels for audio queries.

\vspace{-2mm}
\paragraph{Image-to-Text Loss}
Similar to $\mathcal{L}_{{a2t}}$, we first obtain the text embeddings $\text{proj}(E_T((t^t_k))$ for all class labels. Then, we apply a sigmoid function to the output segmentation map of our model, and use the result to highlight the prediction region on the original image $I_i$. In this process, we retain the foreground information and apply Gaussian blur to the background, which helps preserve the environmental context without losing too much detail. We then pass the highlighted image $I^p_i$ through ImageBind's image encoder $E_I$ to get $t_{v_i}$and utilize projection layer to extract the corresponding visual token $\text{proj}(t_{v_i})$.
The image-to-text loss $\mathcal{L}_{{i2t}}$ can be computed in a similar way:
\begin{equation} \label{i2t}
\begin{split}
T_v &= [\text{proj}(t_{v_0}),\text{proj}(t_{v_1}),...\text{proj}(t_{v_{N-1}})],\\
i2t &= T_v  \times T^t,\\
\mathcal{L}_{{i2t}} &= \text{CELoss}(i2t, y).
\end{split}
\end{equation}
Here, $i2t \in \mathbb{R}^{N \times P}$ represents the similarity scores between the visual features and the text embeddings, and $y \in \mathbb{R}^{N}$ denotes the corresponding index vector of the matched labels.

\vspace{-2mm}
\subsection{Overall Training Loss}
The final mask loss in our model is divided into two parts. For the object-level loss $\mathcal{L}_{sep}$, we calculate the corresponding prediction mask and compare it with the ground truth using Binary Cross-Entropy loss and IoU loss over $N$ masks. For the binary loss across the entire image $\mathcal{L}_{binary}$, we compute a similar loss applied to a single total binary mask for each image.
The overall training loss is expressed as follows:
\begin{equation} \label{total}
\mathcal{L} = \mathcal{L}_{{a2pt}} + \mathcal{L}_{{a2t}} + \mathcal{L}_{{i2t}} + \sum_{i=0}^{N} \mathcal{L}_{sep} + \mathcal{L}_{binary}.
\end{equation}

\section{Experiment}

\subsection{Datasets and Evaluation Metrics}
\paragraph{Datasets.} We evaluate our model on two AVSBench datasets: \textbf{AVSBench-object} \cite{zhou2022audio} and \textbf{AVSBench-semantic} \cite{zhou2024audio}. The \colorbox{gray!20}{AVSBench-object} dataset comprises two scenarios: single sound source segmentation (S4) with 4,932 videos and multiple sound source segmentation (MS3) with 424 videos. MS3 contains 5 annotated frames per video, while S4 requires semi-supervised training, as only the first frame is annotated out of 5 frames.
The \colorbox{gray!20}{AVSBench-semantic} dataset introduces two key differences compared with AVSBench-object. First, it adds a new scenario (V2) containing 6,000 videos with 10 frames per audio clip. Second, it enhances the S4 and MS3 subsets with object-level class label, creating new datasets V1S and V1M, respectively.
\vspace{-3mm}
\paragraph{Metrics.} We adopt two segmentation metrics: $\mathcal{M}_{\mathcal{J}}$ which measures the overlap between ground-truth masks and predicted ones, and $\mathcal{M}_{\mathcal{F}}$ which considers both precision and recall in the evaluation. Following previous work \cite{yang2024cooperation, li2023catr,zhou2022audio,liu2024audio}, we use $\mathcal{M}_{\mathcal{J}}$ and $\mathcal{M}_{\mathcal{F}}$ to denote the mean metrics across the entire dataset. On S4 and MS3, we evaluate binary segmentation performance while on AVSS, we report results based on semantic-level metric $\mathcal{M}^{I}_{\mathcal{J}}$. 

\begin{table}[t]
  \centering
  \footnotesize
  \renewcommand{\arraystretch}{1.0}
\setlength\tabcolsep{4.5mm}
 \begin{tabular}{c|cc|cc}
  \toprule[1pt]
  \multicolumn{1}{c|}{\multirow{2}{*}{Settings}}  & \multicolumn{2}{c|}{\textbf{S4}}  & \multicolumn{2}{c}{\textbf{MS3}}\\
   {} & $\mathcal{M}_{\mathcal{J}}$ & $\mathcal{M}_{\mathcal{F}}$  & $\mathcal{M}_{\mathcal{J}}$ & $\mathcal{M}_{\mathcal{F}}$ \\ \hline
   w/o IBQD &79.6 &0.878 &58.5 &0.664\\
   w/o TbAS &83.6 &0.902 &64.9 &0.716\\
   w/o TbHP &83.9 &0.905 &65.1 &0.713\\
   \rowcolor{c1!50} \textbf{TAViS}  &{84.8} &{0.912} &{68.2} &{0.759}\\

   \bottomrule[1pt]
   \end{tabular}
   \vspace{-3mm}
    \caption{\textbf{Ablation studies of key designs of our TAViS.} Key components of our model are progressively removed to evaluate their individual impact. ``TbHP'' denotes the text-bridged hybrid prompting and ``TbAS'' represents the text-bridged alignment supervision. All ablation studies are conducted with \textbf{224×224} image size.}
  \label{allablation}
 \end{table}

\begin{table}[t]
  \centering
  \footnotesize
  \renewcommand{\arraystretch}{1.0}
\setlength\tabcolsep{3.5mm}
 \begin{tabular}{c|cc|cc}
  \toprule[1pt]
  \multicolumn{1}{c|}{\multirow{2}{*}{Settings}}  & \multicolumn{2}{c|}{\textbf{S4}}  & \multicolumn{2}{c}{\textbf{MS3}}\\
   {} & $\mathcal{M}_{\mathcal{J}}$ & $\mathcal{M}_{\mathcal{F}}$  & $\mathcal{M}_{\mathcal{J}}$ & $\mathcal{M}_{\mathcal{F}}$  \\ \hline
   w/o $\mathcal{L}_{a2t}$ &83.8 &0.905 &65.2 &0.725\\
   w/o $\mathcal{L}_{i2t}$ &84.0 &0.904 &64.4 &0.700\\
   $\mathcal{L}_{a2t}$ +$\mathcal{L}_{i2t}$ + $\mathcal{L}_{a2i}$ &83.6 &0.903 &66.5 &0.744 \\
   $\mathcal{L}_{a2i}$ &83.1 &0.895 &66.5 &0.733 \\
   \rowcolor{c1!50} \textbf{$\mathcal{L}_{a2t}$ +$\mathcal{L}_{i2t}$} &{84.8} &{0.912} &{68.2} &{0.759} \\
   \bottomrule[1pt]
   \end{tabular}
   \vspace{-2mm}
    \caption{\textbf{Ablation studies of our TAViS for text-bridged alignment supervision design (TbAS).} The result of the final line represents the final design of our model, and thus yields the same outcome as the final TAViS result.}
    \vspace{-5mm}
  \label{AVTloss}
 \end{table}

 \begin{table*}[t]
  \centering
  \footnotesize
  \renewcommand{\arraystretch}{1.0}
\setlength\tabcolsep{6.5mm}
 \begin{tabular}{l|c|c|cc|cc|cc}
  \toprule[1pt]
  \multicolumn{1}{l|}{\multirow{2}{*}{Models}} &\multicolumn{1}{c|}{\multirow{2}{*}{Backbone}} &\multicolumn{1}{c|}{\multirow{2}{*}{Size}} & \multicolumn{2}{c|}{\textbf{S4}}  & \multicolumn{2}{c|}{\textbf{MS3}}  & \multicolumn{2}{c}{\textbf{AVSS}}\\
   {} & {} & {} & $\mathcal{M}_{\mathcal{J}}$ & $\mathcal{M}_{\mathcal{F}}$  & $\mathcal{M}_{\mathcal{J}}$ & $\mathcal{M}_{\mathcal{F}}$  & $\mathcal{M}^{I}_{\mathcal{J}}$ \\ \hline
   AVSBench \cite{zhou2023audio} &PVT-v2 &224  &78.7 &0.879 &54.0 &0.645  &- \\
   CATR \cite{li2023catr} &PVT-v2 &224  &81.4 &0.896 &59.0 &0.700 &- \\
   DiffusionAVS \cite{mao2023contrastive} &PVT-v2 &224  &81.4 &0.902 &58.2 &0.709 &- \\
   ECMVAE \cite{mao2023multimodal} &PVT-v2 &224  &81.7 &0.901 &57.8 &0.708 &- \\
   BAVS \cite{liu2024bavs} &PVT-v2 &224   &82.0 &0.886 &58.6 &0.655 &32.6 \\
   AVSegFormer \cite{gao2024avsegformer} &PVT-v2 &224  &82.1 &0.899 &58.4 &0.693 &36.7\\
   UFE-AVS \cite{liu2024audio} &PVT-v2 &224  &83.2 &0.904 &59.5 &0.676 &-\\
   QDFormer \cite{li2024qdformer} &Swin-T &224  &79.5 &0.882 &61.9 &0.661 &-  \\
   COMBO \cite{yang2024cooperation} &PVT-v2 &224 &{84.7} &\bes{0.919} &59.2 &0.712 &\bes{42.1} \\
   SAM \cite{kirillov2023segment} &ViT-H &1024 &55.1 &0.739 &54.0 &0.638 &- \\
   AV-SAM \cite{mo2023av} &ViT-H &1024 &40.5 &0.566 &- &- &- \\
   GAVS \cite{wang2024prompting} &ViT-B &1024 &80.1 &0.902 &63.7 &\bes{0.774} &- \\
    SAMA-AVS \cite{liu2024annotation} &ViT-H &1024 &83.2 &0.901 &66.9 &0.754 &- \\
   \rowcolor{c1!50} \textbf{TAViS} &{ViT-L} &{224} &\bes{84.8} &0.912 &\bes{68.2} &0.759 &\textbf{44.2} \\
   \rowcolor{c1!50} \textbf{TAViS} &\textbf{ViT-L} &\textbf{1024} &\textbf{87.0} &\textbf{0.926} &\textbf{71.2} &\textbf{0.796} &- \\
   
   \bottomrule[1pt]
   \end{tabular}
   \vspace{-3mm}
   \caption{\textbf{Quantitative comparison of our TAViS with other AVS methods on three benchmark datasets.}
`-' indicates the code is not available and that our metric definition differs. As SAM-based methods do not calculate the semantic-level miou, here we do not list the result. Due to computational resource constraints, we do not report results with 1024 image size on the AVSS dataset, which has 10 frames per video. The best and second-best performances under all settings are \textbf{bolded} and \bes{bolded}, respectively.. 
}
\vspace{-5mm}
  \label{SOTA}
 \end{table*}

\subsection{Implementation Details}
We utilize the ViT-L backbone for SAM2 and freeze the parameters of ImageBind and the SAM2 image encoder, while fine-tuning the memory attention and mask decoder. All input frames are resized to 224×224 or 1024×1024. Training is performed using the Adam optimizer \cite{kingma2014adam}, starting with an initial learning rate of 1e-4 and a cosine decay scheduler. The batch sizes used are 4 for the MS3 and AVSS datasets, and 10 for the S4 dataset. Training lasts for 80 epochs on both the MS3 and S4 datasets, and 40 epochs on the AVSS dataset.


\subsection{Ablation Study}

 \begin{table}[t]
  \centering
  \footnotesize
  \renewcommand{\arraystretch}{1.0}
\setlength\tabcolsep{3.8mm}
 \begin{tabular}{c|cc|cc}
  \toprule[1pt]
  \multicolumn{1}{c|}{\multirow{2}{*}{Settings}}  & \multicolumn{2}{c|}{\textbf{S4}}  & \multicolumn{2}{c}{\textbf{MS3}}\\
   {} & $\mathcal{M}_{\mathcal{J}}$ & $\mathcal{M}_{\mathcal{F}}$  & $\mathcal{M}_{\mathcal{J}}$ & $\mathcal{M}_{\mathcal{F}}$  \\ \hline
   w/o $\bm{p}^t$ &83.6 &0.903 &63.4 &0.714\\
   MLP prediction &84.1 &0.904 &63.6 &0.710 \\
   w/o $\bm{t}_{v}$ &83.3 &0.902 &63.6 &0.701\\
   \rowcolor{c1!50} \textbf{$\bm{p}^t$+$\bm{p}^a$+$\bm{t}_{v}$} &{84.8} &{0.912} &{68.2} &{0.759}  \\
   \bottomrule[1pt]
   \end{tabular}
   \vspace{-2mm}
    \caption{\textbf{Ablation studies of our TAViS for text-bridged hybrid prompting (TbHP).}  }
   \vspace{-4mm}
  \label{prompt}
 \end{table}

  \begin{table}[t]
  \centering
  \footnotesize
  \renewcommand{\arraystretch}{1.0}
\setlength\tabcolsep{1.3mm}
 \begin{tabular}{c|c|c|c|cc|cc}
  \toprule[1pt]
  \multicolumn{1}{c|}{\multirow{2}{*}{Settings}} &\multicolumn{1}{c|}{\multirow{2}{*}{Model}} &\multicolumn{1}{c|}{\multirow{2}{*}{MACs}} &\multicolumn{1}{c|}{\multirow{2}{*}{Size}} & \multicolumn{2}{c|}{\textbf{S4}}  & \multicolumn{2}{c}{\textbf{MS3}}\\
   {} & {} &{} &{} & $\mathcal{M}_{\mathcal{J}}$ & $\mathcal{M}_{\mathcal{F}}$  & $\mathcal{M}_{\mathcal{J}}$ & $\mathcal{M}_{\mathcal{F}}$  \\ \hline
   SAMA-AVS &SAM-H &598G &1024 &83.2 &0.901 &66.9 &0.754 \\
   \rowcolor{c1!50} \textbf{TAViS} &SAM-H &334G &224 &84.3 &0.911 &67.3 &0.743\\\hline
   SAMA-AVS &SAM2-L &221G &224 &83.5 &0.902 &66.3 &0.727\\
   \rowcolor{c1!50} \textbf{TAViS} &SAM2-L &255G &224  &{84.8} &{0.912} &{68.2} &{0.759}  \\
   \bottomrule[1pt]
   \end{tabular}
   \vspace{-2mm}
    \caption{\textbf{A fair comparison with SAM-based SOTA methods.}  }
   \vspace{-4mm}
  \label{differentfoundation}
 \end{table}
 
\paragraph{Architecture Design.}
To demonstrate the efficacy of various components in our model, we conduct ablation studies and report the results in Table~\ref{allablation}. We progressively remove key components from our model to analyze their contribution. First, removing the IBQD module leads to significant performance degradation across all scenarios, demonstrating the importance of decomposing audio queries for precise object-level audio-visual alignment. Subsequently, removing the text-bridged alignment supervision (TbAS) results in notable performance drops, highlighting its effectiveness for accurate object localization. Finally, the elimination of the text-bridged hybrid prompting (TbHP) also leads to performance decline, validating the benefits of our design.

\vspace{-4mm}
\paragraph{Text-bridged Alignment Supervision.}
As shown in Table~\ref{AVTloss}, we conduct ablation studies on our text-bridged alignment supervision, \ie, $\mathcal{L}_{a2t}$ and $\mathcal{L}_{i2t}$ across two datasets. Removing $\mathcal{L}_{a2t}$ and $\mathcal{L}_{i2t}$ separately leads to performance degradation, highlighting the importance of both audio-to-text and visual-to-text alignments. 
To further investigate the role of text embedding, we introduce an additional audio-visual alignment loss $\mathcal{L}_{a2i}$, which aligns the $\text{proj}(t'_{a_i})$ and $\text{proj}(t_{v_i})$ on top of our baseline design. However, incorporating this loss harms performance.
We hypothesize that this is due to the noisy nature of the audio and visual tokens, which are generated from audio queries and segmentation masks without accurate ground truth supervision. Aligning these noisy tokens likely introduces uncertainty, degrading performance. This hypothesis is further supported by experiments where we solely use $\mathcal{L}_{a2i}$, which results in reduced performance.

\begin{figure}[!t]
    \centering
    \includegraphics[width=1\linewidth]{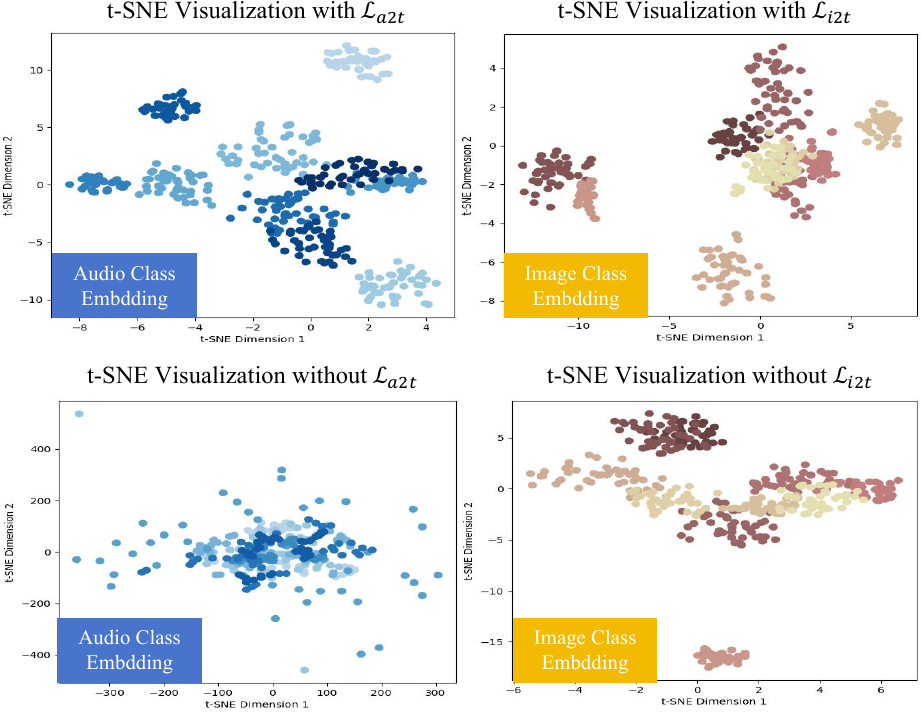}
    \vspace{-5mm}
    \caption{\textbf{t-SNE visualization with text-bridge (up) and without text-bridge (down).} Different colors indicate different classes.}
    \label{tsne}
    \vspace{-4mm}
\end{figure}

To further demonstrate the effectiveness of using text as a bridge, we employ t-SNE \cite{van2008visualizing} to visualize the audio class and visual class embeddings under different settings. For clearer visualization, we focus on the S4 dataset results, which contains single-class images. We specifically exclude classes with insufficient samples, as it hard to cluster. As illustrated in Figure~\ref{tsne}, utilizing text as an intermediate bridge leads to more compact clusters within each class, effectively mitigating the impact of modality-specific noise. In contrast, when removing either text-related loss term ($\mathcal{L}_{a2t}$ or $\mathcal{L}_{i2t}$), the direct audio-visual alignment results in more dispersed clusters, indicating the challenge of handling noisy representations without textual guidance.

\vspace{-3mm}
\paragraph{Text-bridged Hybrid Prompting.}
Moreover, we investigate the influence of our text-bridged hybrid prompting, which consists of two components: sparse prompt and dense prompt. We first evaluate the impact of sparse prompt by removing the text prompt $\bm{p}^t$ while only maintaining the audio prompt $\bm{p}^a$. As shown in Table~\ref{prompt}, this ablation leads to performance degradation, highlighting the crucial role of text-based prototype information in segmentation tasks.
Furthermore, we experiment with an alternative text generation approach by removing the text encoder and directly predicting pseudo-text embedding via MLP.
This modification results in a significant performance drop on MS3. We conjecture this is because text prediction is highly overfitted on small-scale datasets without the pre-trained knowledge in the text encoder.
Similarly, removing the dense prompt $\bm{t}_{v}$ also degrades performance, particularly for $\mathcal{M}_{\mathcal{F}}$, demonstrating its vital role in capturing fine-grained segmentation details.

\begin{figure*}[!t]
    \centering
    \includegraphics[width=1\linewidth]{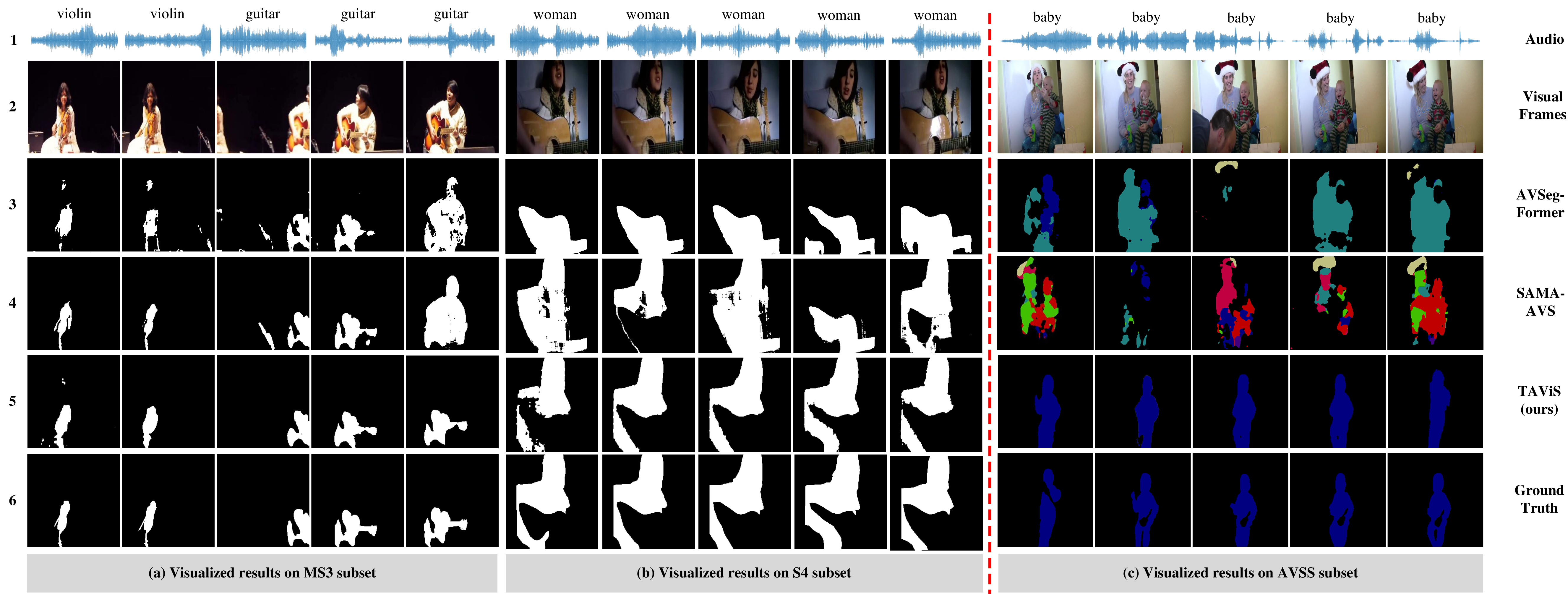}
    \vspace{-5mm}
    \caption{\textbf{Qualitative comparison of our model against state-of-the-art AVSS methods.} From top to bottom: (1) audio visibility graphs; (2) visual frames; (3) prediction maps from AVSegFormer \cite{gao2024avsegformer}; (4) prediction maps from SAMA-AVS \cite{liu2024annotation} (left) and AVSBench \cite{zhou2023audio} (right); (5) prediction maps from our model; and (6) ground truth.  }
    \label{compare_sota}
    \vspace{-4mm}
\end{figure*}


\subsection{Comparison with State-of-the-Art Methods}
We report the performance comparison of our methods against top-performing state-of-the-art methods, including 14 AVS methods \cite{zhou2023audio, hao2024improving, li2023catr, mao2023contrastive, mao2023multimodal, liu2024bavs, gao2024avsegformer, liu2024audio, li2024qdformer, yang2024cooperation, kirillov2023segment, mo2023av, liu2024annotation, wang2024prompting}. As shown in Table~\ref{SOTA}, we first compare performance on the S4 and MS3 datasets, and our model demonstrates consistent improvements over existing methods on both datasets.
Furthermore, we evaluate our model on the AVSS dataset by computing the similarity between audio embedding $T_a$ and text embedding $T^t$ for class prediction of segmented regions. Our approach not only surpasses existing methods on AVSS but also extends SAM-based methods with class prediction capabilities. 
Notably, increasing the image size from 224 to 1024 significantly improves performance, highlighting the importance of image resolution in SAM(SAM2)-based models, as 1024 matches the foundation model’s original training resolution. 
Figure~\ref{compare_sota} provides visual comparisons among the top-performing models.

To ensure a fair comparison with previous SAM-based methods, we change our SAM2 foundation model to SAM, as shown in Table~\ref{differentfoundation}. 
Even though with small image size, our model still outperforms the SAM-based SAMA-AVS, demonstrating the effectiveness of our design. Furthermore, we replaced the SAM-based model with SAM2 and retrained the model following the procedure described in the paper. With SAM2, SAMA-AVS showed only a slight improvement, while still performing below our model. These comparative experiments demonstrate that our superior performance is not solely attributed to the use of SAM2, but also stems from our novel model design, which provides greater advantages than the additional pre-training dataset introduced in SAMA-AVS.
As our SAM2 version achieves better performance with lower computation cost, we ultimately select SAM2 as our segmentation foundation model.


\subsection{Analysis of Zero-shot Generalization Ability}
\label{zero-shot}
Previous AVSS methods typically predict classes by applying an MLP operation on a fixed set of classes \cite{yang2024cooperation, mo2024unveiling}, which limits their ability to generalize in real-world scenarios. In contrast, our model leverages ImageBind, which was trained on a broader set of datasets that extend beyond single audio-visual datasets. This enables superior generalization to unseen classes. Following the approach and setting in \cite{guo2024open}, we train the model on seen classes and evaluate it on unseen classes. 
To predict the specific class, we compute the similarity of text embeddings with audio and image embeddings respectively, combine them with an addition operation (coefficient set to 1) to obtain the final class prediction.
As shown in Table~\ref{ovavss}, our model outperforms existing OV-AVSS methods 
with less trainable parameters  on zero-shot setting, highlighting the effectiveness of aligning foundation models. 
While more specialized AVS foundation models may be better suited for the AVSS task \cite{mo2024unveiling}, the lack of text alignment in such models hinders their generalization ability.

 \begin{table}[t]
  \centering
  \footnotesize
  \renewcommand{\arraystretch}{1}
\setlength\tabcolsep{4.5mm}
 \begin{tabular}{c|c|c}
  \toprule[1pt]
  \multicolumn{1}{c|}{\multirow{1}{*}{Settings}}  & {\textbf{Zero-shot $\mathcal{M}^{I}_{\mathcal{J}}$}}& {\textbf{Trainable Param}}\\
\hline
   OV-AVSS \cite{guo2024open} &22.20 &183.6M\\
   \rowcolor{c1!50} TAViS &28.21 &54.9M\\
   \bottomrule[1pt]
   \end{tabular}
   \vspace{-3mm}
    \caption{\textbf{Comparison with one zero-shot audio-visual segmentation method on AVSBench-OV dataset.}   }
   \vspace{-6mm}
  \label{ovavss}
 \end{table}

\section{Conclusion}
In this paper, we propose TAViS, a novel framework that effectively integrates ImageBind and SAM2 to address the AVS task. First, we design IBQD module that effectively separates audio sources to adapt SAM2 for object-level segmentation.
To bridge the two foundation models, we develop a text-bridged hybrid prompting technique and introduce text-bridged alignment supervision through audio-to-text and image-to-text losses.
Our model demonstrates superior performance across diverse AVS benchmarks while excelling in zero-shot settings.

{\small
\bibliographystyle{ieee_fullname}
\bibliography{main_iccv}
}

\end{document}